%% file: all.tex
\documentclass[runningheads]{llncs}
\usepackage{eccv}
\usepackage{graphicx}
\usepackage{booktabs}
\usepackage{amsmath,amssymb}
\usepackage{eccvabbrv}
\usepackage{multirow}
\usepackage[accsupp]{axessibility}
\usepackage[pagebackref]{hyperref}
\usepackage[table]{xcolor}
\usepackage{orcidlink}

\newcommand{\best}[1]{\textbf{#1}}
\newcommand{\second}[1]{\underline{#1}}

\begin{document}

\input{main2all}

\clearpage
\appendix
\section*{Supplementary Material}
\input{supp2all}

\bibliographystyle{splncs04}
\bibliography{main}

\end{document}

%% file: main2all.tex
\graphicspath{{./Figs/}}

\title{Towards Joint Quantization and Token Pruning of Vision-Language Models} 

\titlerunning{Joint Quantization and Token Pruning for VLMs}

\author{Xinqing Li\inst{1}\orcidlink{0009-0001-0293-4144} \and
Xin He\inst{2}\orcidlink{0009-0004-2139-6590} \and
Xindong Zhang\inst{3}\orcidlink{0000-0001-6732-514X} \and
Ming-Ming Cheng\inst{1,4,5}\orcidlink{0000-0001-5550-8758} \and
Lei Zhang\inst{3,6}\orcidlink{0000-0002-2078-4215} \and
Yun Liu\inst{1,4,5}\thanks{Corresponding author: Yun Liu (liuyun@nankai.edu.cn)}\orcidlink{0000-0001-6143-0264}}

\authorrunning{X.~Li et al.}

\institute{VCIP, College of Computer Science, Nankai University \and
School of Computer Science and Engineering, Tianjin University of Technology \and
OPPO Research Institute \and
Academy for Advanced Interdisciplinary Studies, Nankai University \and
Nankai International Advanced Research Institute, Shenzhen Futian \and
Department of Computing, Hong Kong Polytechnic University
}
\maketitle


\begin{abstract}
Deploying Vision-Language Models (VLMs) under aggressive low-bit inference remains challenging because inference cost is dominated by the long visual-token prefix during prefill and the growing KV cache during autoregressive decoding. Token pruning and low-bit quantization are complementary for reducing these costs, yet naive stage-wise combinations are often brittle due to a mismatch between quantization calibration and pruning execution. We present a collaborative quantization-and-pruning framework that unifies low-bit inference and deterministic visual-token pruning in a single deployable pipeline. The framework introduces the \textbf{Q}uantization \textbf{U}nified \textbf{O}ffline \textbf{T}oken \textbf{A}llocator (\textbf{QUOTA}), which converts low-bit calibration signals into a layer-wise token allocation schedule and materializes it as a pruning recipe. Token importance is evaluated under deployed W4A4 operators with a quantized KV cache by combining activation magnitude, attention cues, and an explicit low-bit risk signal, enabling consistent budgeted top-$k$ selection. Experiments on standard VLM benchmarks show improved robustness over stage-wise baselines under the same low-bit regime, achieving 95.65\% average retention while retaining only 30\% of visual tokens, compared with about 94.3\% retention for representative stage-wise combinations. The code will be released.
  \keywords{Vision-Language Models \and Quantization \and Token Pruning \and Joint Quantization and Token Pruning}
\end{abstract}

\section{Introduction}
\label{sec:intro}
Vision-Language Models (VLMs) have emerged as a dominant backbone for multimodal understanding and reasoning, yet deployment remains challenging due to high inference cost and memory footprint~\cite{lin2024vila, shang2025llava}. Two practical bottlenecks dominate end-to-end latency and memory traffic. During prefill, the vision encoder and multimodal connector produce a long visual-token prefix that accounts for most of the compute. During autoregressive decoding, the continuously growing KV cache becomes a bandwidth bottleneck and inflates memory pressure~\cite{huang2025dynamic, yang2025topv}. These issues are particularly severe on resource-constrained GPUs and edge servers, motivating compression methods that reduce both prefill compute and decoding-time KV cache cost in a directly deployable manner. A practical solution should address these two bottlenecks jointly under the same low-bit inference regime used at deployment.

\begin{figure*}[t]
  \centering
  \includegraphics[width=\textwidth]{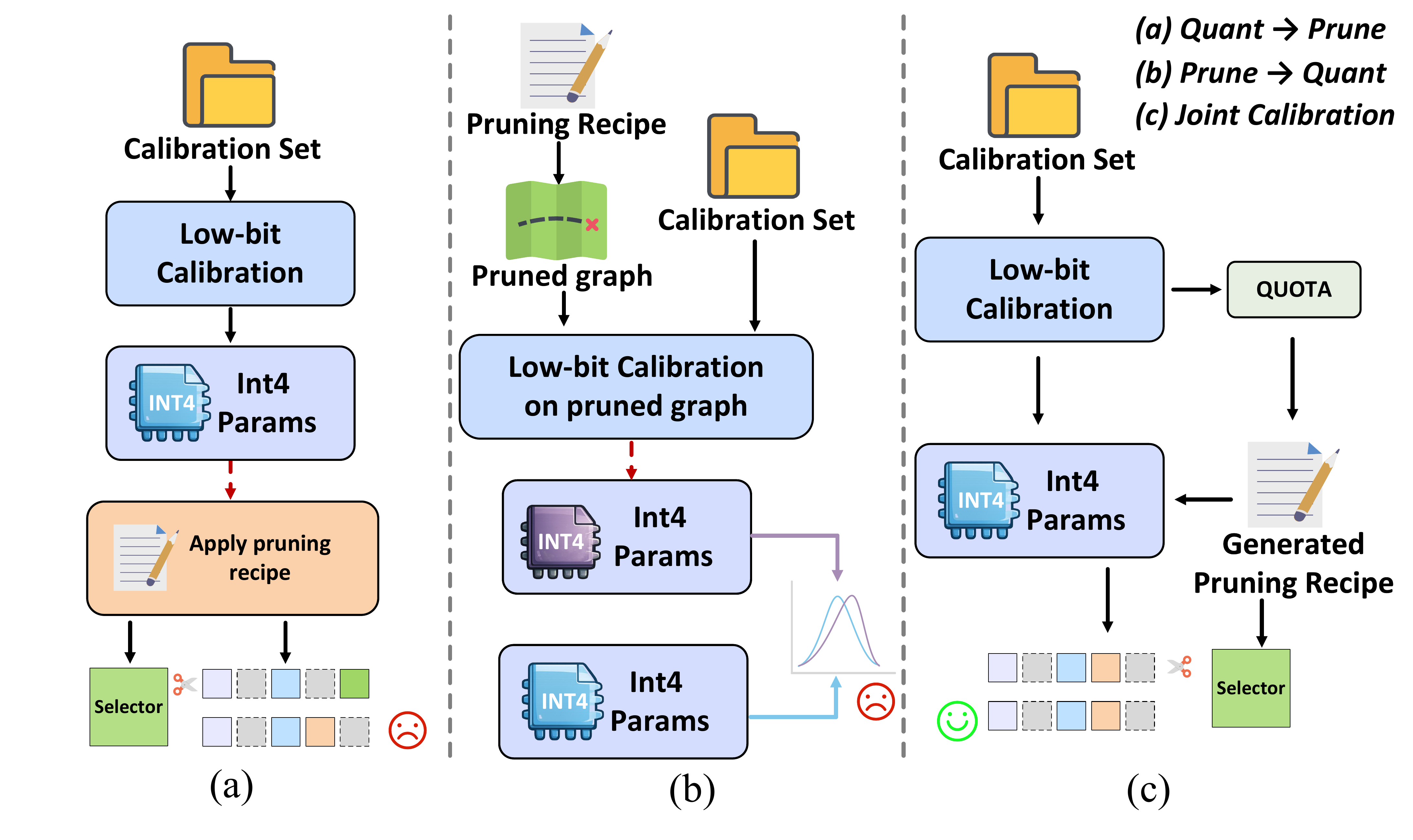}
  \caption{Stage-wise \vs unified pipelines for low-bit calibration and visual-token pruning.}
  \label{fig:pipeline}
\end{figure*}

Existing acceleration efforts for VLMs have largely progressed along two relatively independent directions. 
Token reduction methods, including token pruning~\cite{yang2025topv,rao2021dynamicvit} and token merging~\cite{bolya2023token,shang2025llava}, shorten the visual-token prefix processed by the language model and reduce cross-modal attention cost, leading to substantial savings in prefill latency. 
Quantization methods compress weights and activations, and in some cases the KV cache, to reduce memory footprint and improve throughput under low-bit arithmetic~\cite{frantar2023gptq,xiao2023smoothquant,hooper2024kvquant}, which is particularly effective at alleviating decoding-time memory-bandwidth pressure. 
Although these directions are complementary, practical deployments often combine them in a stage-wise manner (\eg, quantize-then-prune or prune-then-quantize)~\cite{han2015deep,kim2026prunethenquantize}. 
As illustrated in Fig.~\ref{fig:pipeline}, such sequential pipelines can introduce a mismatch between the low-bit calibration manifold and the pruned execution form, making the combined system brittle under aggressive low-bit regimes. 
This motivates a unified deployment-time pipeline that aligns token pruning with the calibrated low-bit operating regime.

To this end, we propose a practical collaborative quantization-and-pruning framework for VLM deployment that executes visual-token pruning \emph{within} the same low-bit inference regime used at deployment. Our pipeline pairs a calibrated W4A4 model with a deterministic layer-wise pruning recipe, and applies pruning directly in the quantized forward pass with a quantized KV cache, ensuring consistent attention masks, positions, and cached keys/values for retained tokens. To instantiate the recipe in a training-free manner, we introduce the \textbf{Q}uantization \textbf{U}nified \textbf{O}ffline \textbf{T}oken \textbf{A}llocator (\textbf{QUOTA}), which converts low-bit calibration sensitivities into a layer-wise token allocation schedule and materializes it as a deployable pruning recipe. As a result, token pruning reduces the visual-token prefix and the associated cross-modal attention and KV cache cost while operating under the calibrated low-bit operators.

Our contributions are threefold:
\begin{itemize}
  \item We introduce QUOTA, a calibration-driven offline token allocator that converts low-bit sensitivity signals into a layer-wise token budget schedule and materializes it as a deployable deterministic pruning recipe.

  \item We study stage-wise combinations of token pruning and low-bit quantization in VLMs, and characterize accuracy degradation caused by the mismatch between low-bit calibration and pruned execution, which motivates a unified collaborative pipeline.

  \item We evaluate the proposed framework on standard VLM benchmarks and deployment-oriented efficiency metrics, demonstrating improved robustness and a better accuracy--efficiency trade-off than representative stage-wise baselines under the same low-bit regime.
\end{itemize}

\section{Related Work}
\label{sec:related}
Quantization is a key technique for reducing memory footprint and improving inference throughput in large models by representing weights and activations in low-bit formats. Two main paradigms are Post-Training Quantization (PTQ)~\cite{frantar2023gptq,xiao2023smoothquant} and Quantization-Aware Training (QAT)~\cite{jacob2018quantization}. PTQ is particularly attractive for large-scale deployment as it requires little to no retraining. Representative methods such as GPTQ~\cite{frantar2023gptq} and SmoothQuant~\cite{xiao2023smoothquant} focus on local error minimization and calibration procedures to reduce degradation under 4-8 bit quantization. 
Recent work extends quantization to the KV cache~\cite{hooper2024kvquant,kwon2023efficient}, which helps alleviate the memory bandwidth bottleneck during autoregressive decoding, a key concern for VLMs. These methods assume a fixed inference graph and relatively stable activation statistics during calibration~\cite{gong2024makes}. However, in practice, token-level dynamics such as pruning can invalidate these assumptions, introducing a mismatch between calibration and deployed execution. In parallel, token reduction methods focus on shrinking the visual prefix to reduce prefill latency and attention cost, which we review next.

Token reduction methods for VLMs mainly fall into two categories: token pruning~\cite{ye2025fit,yang2025topv,zhang2025beyond} and token merging~\cite{bolya2023token,shang2025llava}. Token pruning selects a subset of informative visual tokens according to importance scores and removes the rest, while token merging shortens the sequence by aggregating similar tokens into fewer representations. Many recent VLM-focused approaches adopt training-free designs and emphasize compatibility with modern attention implementations, reporting substantial prefill acceleration by shortening the visual-token prefix and reducing cross-modal attention cost~\cite{yang2025topv,zhang2025sparsevlm,li2025balanced,shang2025llava}. 
However, most token reduction methods were evaluated in full precision or under mild quantization, and they rarely examined the reliability of token-importance estimation under aggressive low-bit inference~\cite{bondarenko2021understanding,gong2024makes}. 
In particular, quantization noise can perturb importance signals, and pruning alters the executed token set and attention patterns, together introducing a mismatch between how importance is calibrated/estimated and how the pruned model is executed. 

The combination of pruning and quantization has been explored in model compression, \eg, developed as stage-wise pipelines~\cite{han2015deep}. Most joint designs couple quantization with structured pruning, such as weight- or channel-level pruning~\cite{guo2024compressing,zeng2025gqsa,mozaffari2025slim,qu2025automatic}, to reduce parameter storage and arithmetic cost under a largely fixed computation graph. Because such structured pruning preserves the token sequence length, it leaves the cross-modal attention complexity and the KV cache growth pattern largely unchanged. In contrast, token pruning acts on the sequence dimension by shortening the visual-token prefix, directly reducing both attention cost and KV cache footprint. This distinction is particularly important for VLMs, where the visual-token prefix can dominate prefill cost and the resulting cache growth amplifies decoding-time memory-bandwidth pressure. As a result, conclusions drawn from structured pruning-quantization co-design do not directly account for the calibration and execution mismatch introduced by token-level pruning under aggressive low-bit inference. 

A small body of recent work has begun to examine the incompatibility between token pruning and aggressive low-bit quantization, and has proposed co-designed pipelines to mitigate it ~\cite{fang2025sqap}. These studies suggest a bidirectional coupling: quantization noise can perturb attention patterns and reduce the reliability of token-importance estimation, while token pruning can reshape activation statistics and invalidate the calibration assumptions underlying low-bit operators. Together, these findings underscore the need for deployment-oriented designs that keep token pruning consistent with the calibrated low-bit operating regime, a setting that remains underexplored for general-purpose VLM inference.

\section{Method}
\label{sec:method}
This section presents a post-training collaborative quantization-and-pruning framework for deploying VLMs under an aggressive low-bit regime, including W4A4 inference with a quantized KV cache. We first define the unified low-bit execution form, and then introduce QUOTA, which transforms low-bit calibration signals into a deterministic layer-wise token budget schedule and instantiates it as a deployable pruning recipe executed within the same low-bit inference pipeline.

\subsection{Framework Overview: Unified Low-bit Inference with Deterministic Token Pruning}
\label{subsec:framework}
\begin{figure}[t]
  \centering
  \includegraphics[width=\linewidth]{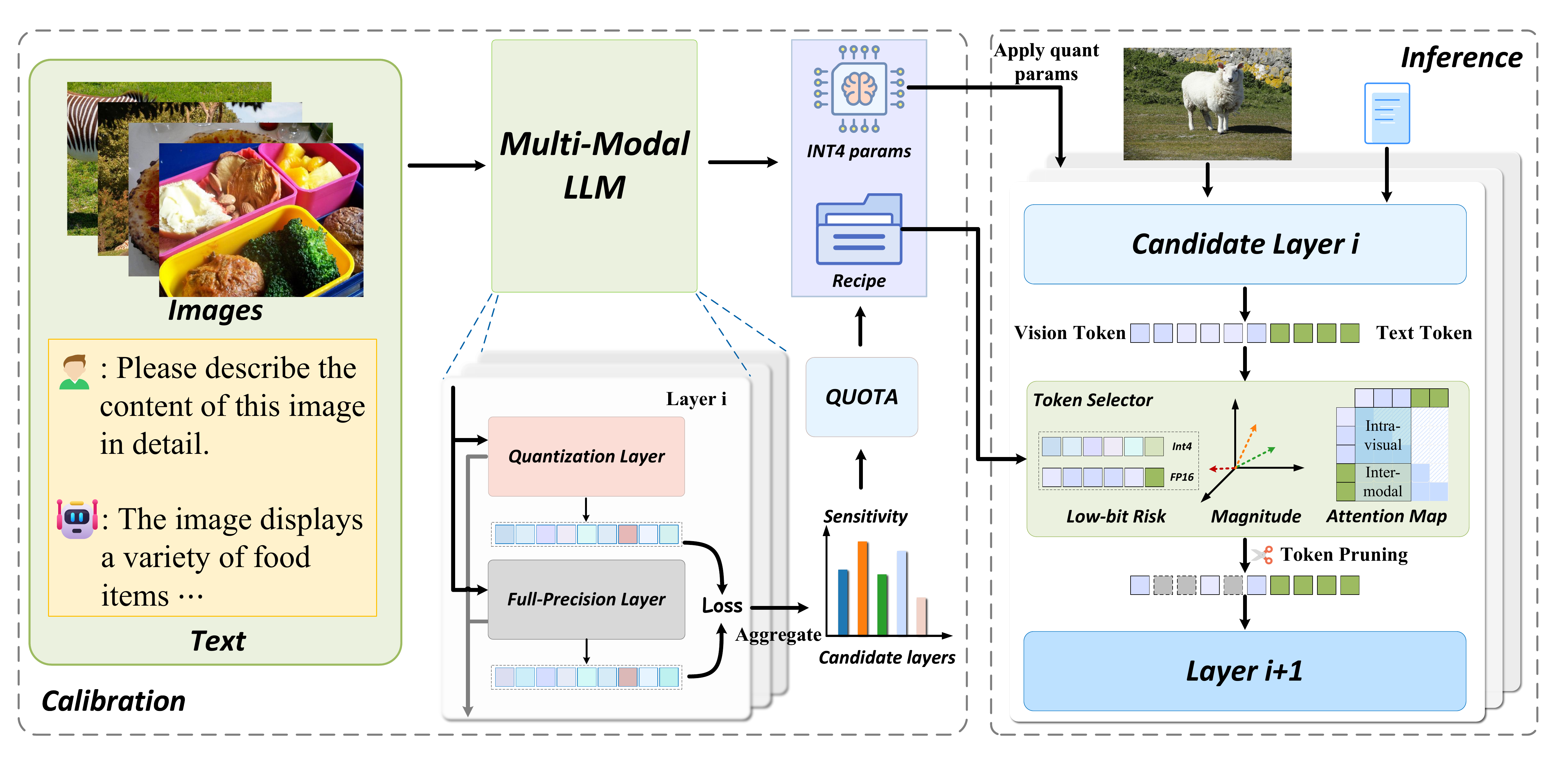}
  \caption{Overview of the collaborative pipeline: QUOTA derives a pruning recipe from low-bit calibration, which is executed during deployment under quantized inference.}
  \label{fig:framework}
\end{figure}
As shown in Fig.~\ref{fig:framework}, our pipeline unifies recipe generation and deployment under the same low-bit operating regime. During calibration, QUOTA converts low-bit sensitivity signals into a deterministic pruning recipe that specifies the candidate layer set and a layer-wise keep-ratio schedule (\ie, token budgets). During deployment, the calibrated W4A4 model executes token pruning inside the quantized forward pass by enforcing each budget with deterministic top-$k$ selection.
Given the layer-wise budgets, token importance is evaluated under the deployed low-bit operators (including a quantized KV cache). In addition to standard cues such as activation magnitude and attention, we optionally incorporate an explicit low-bit risk signal to improve the consistency of token selection under quantized execution.
At inference time, we align token pruning with the calibrated low-bit regime by maintaining consistent alignment among token indices, positional indices, attention masks, and the KV cache, so that keys and values are materialized and cached only for retained tokens. By avoiding stale cached states and position mask mismatches, this design preserves stable autoregressive decoding under KV cache quantization.

\subsection{Sensitivity-Conditioned Token Budgets}
\label{subsec:budgets}
We next derive the layer-wise token budgets used by the deterministic pruning recipe.  Prior works show that the effect of token pruning is strongly layer dependent, and layer-wise schedules are important for balancing accuracy and efficiency~\cite{rao2021dynamicvit,yang2025topv,li2025balanced}. Because pruning shortens the token sequence processed by subsequent blocks, earlier pruning yields larger cumulative savings in cross-modal attention computation and KV cache growth. However, in VLMs, the language model exhibits a depth-wise transition in how it incorporates visual tokens, making pruning too early unreliable and pruning too late overly sensitive~\cite{wu2026vision}. We therefore select a contiguous early layer range as pruning candidates and keep the final blocks conservative, and then condition the per-layer budgets on low-bit calibration sensitivity to obtain a deployable keep-ratio schedule.

\begin{figure}[t]
  \centering
  \includegraphics[width=\linewidth]{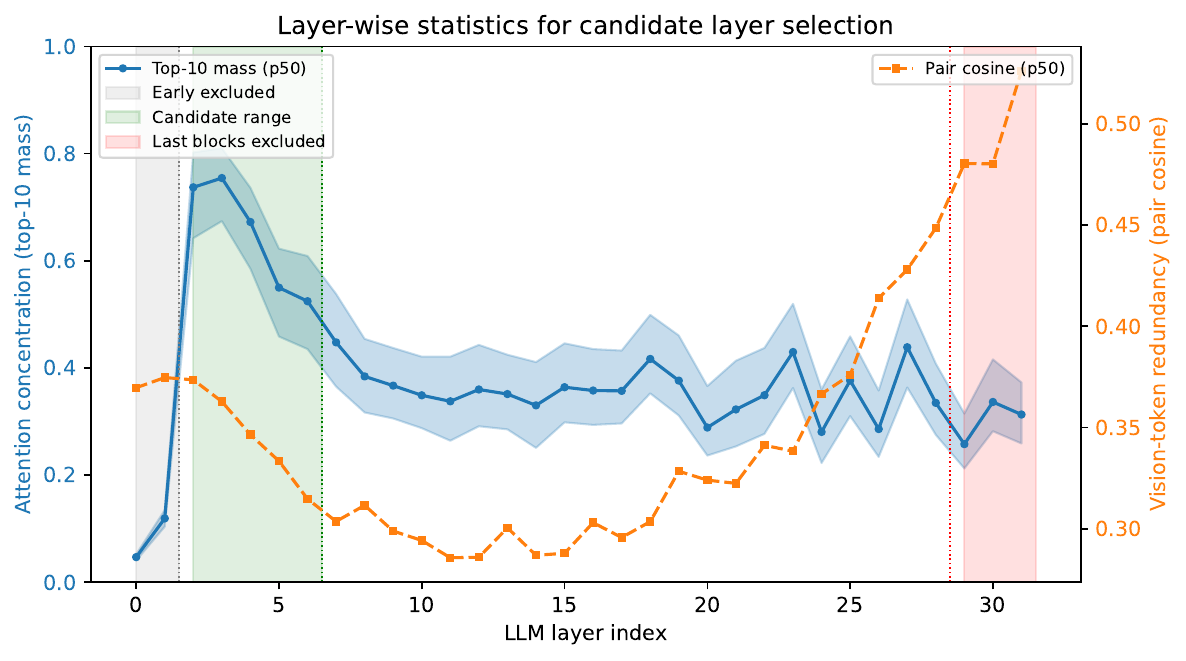}
  \caption{Layer-wise attention concentration and visual-token redundancy used to define the candidate layer set $\mathcal{L}_c$.}
  \label{fig:layer_select}
\end{figure}

To instantiate the candidate layer set $\mathcal{L}_c$ in the pruning recipe, we profile layer-wise attention concentration and visual-token redundancy on a calibration set, as shown in Fig.~\ref{fig:layer_select}. Attention concentration over visual tokens is measured from the inter-modal attention block of the multimodal attention map, using the median top-10 attention mass from text queries to visual tokens together with the interquartile range (IQR) across samples. The concentration is low in the first two layers and increases sharply afterwards, indicating the earliest depth at which attention-based importance ranking becomes more reliable. In parallel, a redundancy proxy based on the median pairwise cosine similarity among visual tokens remains low in the early blocks and is substantially lower than in the final blocks, suggesting that visual tokens are still diverse at this stage. Based on these trends, we exclude the first two layers, select a contiguous early-layer range as $\mathcal{L}_c$, and avoid pruning in the final blocks, where cumulative savings are smaller and token removal is more fragile. The resulting $\mathcal{L}_c$ is used in the subsequent budget allocation and token selection steps.

Given $\mathcal{L}_c$, QUOTA derives layer-wise token budgets during low-bit calibration by profiling quantization sensitivity on a calibration set. For each $\ell \in \mathcal{L}_c$, we measure sensitivity as the relative deviation between a full-precision activation $x_{\ell}$ and its low-bit counterpart $x_{\ell}^{q}$ computed under the deployed low-bit operators:
\begin{equation}
S_\ell \;=\; \mathrm{median}_{x\sim \mathcal{D}_{\mathrm{cal}}}\left[
\frac{\left\|x^q_\ell - x_\ell\right\|_2}{\left\|x_\ell\right\|_2+\eta}
\right],\quad \ell\in\mathcal{L}_c,
\label{eq:qsens}
\end{equation}
where $\eta$ is a small constant for numerical stability. A larger $S_\ell$ indicates that layer $\ell$ is more sensitive under low-bit inference; QUOTA therefore allocates a larger token budget to that layer.

Raw sensitivities can be heavy-tailed across layers, so we apply percentile-based clipping and normalization. Let $P_{10}$ and $P_{90}$ denote the 10th and 90th percentiles of $\{S_\ell\}_{\ell\in\mathcal{L}_c}$. We compute
\begin{equation}
\tilde{S}_\ell \;=\;
\mathrm{clip}\!\left(\frac{S_\ell - P_{10}}{P_{90}-P_{10}},\, 0.1,\, 0.9\right),\quad \ell\in\mathcal{L}_c,
\label{eq:norm_sens}
\end{equation}
where we rescale sensitivities by the central percentile range and cap extreme layers to prevent unstable budget allocation. QUOTA maps ${\tilde{S}_\ell}$ to a layer-wise keep-ratio schedule by distributing a global drop budget over the ordered candidate layers $\mathcal{L}_c={\ell_1,\ldots,\ell_m}$. We assign larger drop to less sensitive layers via a temperature-controlled softmax:
\begin{equation}
\pi_i \;=\; 
\frac{\exp\!\left(\frac{1-\tilde{S}_{\ell_i}}{\tau}\right)}
{\sum_{j=1}^{m}\exp\!\left(\frac{1-\tilde{S}_{\ell_j}}{\tau}\right)},
\quad i=1,\ldots,m .
\label{eq:layer_share}
\end{equation}
where $\tau$ controls the sharpness. To ensure robustness and a monotone schedule, we enforce a keep-ratio floor $p_{\min}$ and set the total drop budget $B=1-p_{\min}$. We assign $d_i=B\pi_i$ and form a non-increasing keep-ratio table:
\begin{equation}
r_{\ell_i} \;=\; \max\!\left(p_{\min},\, 1-\sum_{j=1}^{i} d_j\right),\quad i=1,\ldots,m.
\label{eq:keep_schedule}
\end{equation}
The resulting keep-ratio schedule is materialized as a deployable pruning recipe and consumed by the inference pipeline.

\subsection{Quantization-Consistent Token Selection}
\label{subsec:score}
Given the layer-wise keep ratios $\{r_\ell\}_{\ell\in\mathcal{L}_c}$, we perform budgeted token selection at each candidate layer. 
We follow a quantization-consistent principle: all importance signals are computed under the deployed low-bit operators, i.e., from the W4A4 forward pass with a quantized KV cache. We then form a deterministic composite score and apply top-$K_\ell$ selection.

At a candidate layer $\ell\in\mathcal{L}_c$, let $V^\ell=\{v_i^\ell\}_{i=1}^{N_\ell}$ and $T^\ell$ denote the visual and text token representations, respectively. Let $V_0$ be the reference visual-token length after the projector, measured during calibration and stored in the pruning recipe. The layer-wise budget is

\begin{equation}
K_\ell=\left\lceil r_\ell\,V_0\right\rceil,
\end{equation}
We retain the top-$K_\ell$ visual tokens ranked by the score defined below. For each visual token $v_i^\ell$, we compute four importance metrics from the deployed quantized forward pass. We use the convention that attention weights $A_{qk}$ are indexed by \emph{query} token $q$ and \emph{key} token $k$. Specifically,
\begin{equation}
\begin{aligned}
&m^{\mathrm{mag}}_{i,\ell} \;=\; \|v_i^\ell\|_2,
\qquad
m^{\mathrm{inter}}_{i,\ell} \;=\; \frac{1}{H}\sum_{h=1}^{H}\sum_{j\in \mathcal{T}} A^{\mathrm{inter},h}_{j i}(\ell), \\
&m^{\mathrm{res}}_{i,\ell} \;=\; \|Q(v_i^\ell)-v_i^\ell\|_2,
\qquad
m^{\mathrm{intra}}_{i,\ell} \;=\; \frac{1}{H}\sum_{h=1}^{H}\sum_{k\in \mathcal{V}} A^{\mathrm{intra},h}_{k i}(\ell),
\end{aligned}
\label{eq:metrics_block}
\end{equation}
where $H$ is the number of heads, $\mathcal{T}$ and $\mathcal{V}$ index text and visual tokens, $A(\cdot)$ is computed by the deployed quantized operators, and $Q$ is the activation quantizer. The residual $m^{\mathrm{res}}_{i,\ell}$ serves as a low-bit risk signal.

The metrics have different scales and can be affected by outliers, so we apply a per-layer robust normalization operator $\mathcal{G}$ based on percentile clipping and rescaling. For each metric type $m\in\{\mathrm{mag},\mathrm{inter},\mathrm{intra},\mathrm{res}\}$, we compute
\begin{equation}
\hat{m}^{(m)}_{i,\ell} \;=\; \mathcal{G}\!\left(m^{(m)}_{i,\ell}\right),
\quad i=1,\ldots,N_\ell,
\label{eq:metric_norm}
\end{equation}
where $\mathcal{G}$ is applied independently at each layer $\ell$ to map values to a comparable range while reducing the influence of extreme tokens. We form a composite importance score using a layer-shared weighted sum of the normalized metrics:
\begin{equation}
\mathrm{score}_{i,\ell} \;=\;
w_m \hat{m}^{(\mathrm{mag})}_{i,\ell}
+ w_c \hat{m}^{(\mathrm{inter})}_{i,\ell}
+ w_s \hat{m}^{(\mathrm{intra})}_{i,\ell}
+ w_r \hat{m}^{(\mathrm{res})}_{i,\ell},
\label{eq:score_fusion}
\end{equation}
where $\{w_m,w_c,w_s,w_r\}$ are shared across layers. We then retain a deterministic set of $K_\ell$ visual tokens at each candidate layer by top-$K_\ell$ selection:
\begin{equation}
\mathcal{I}_\ell \;=\; \mathrm{TopK}\Big(\{\mathrm{score}_{i,\ell}\}_{i\in\mathcal{V}},\; K_\ell\Big),
\quad \ell\in\mathcal{L}_c,
\label{eq:topk}
\end{equation}
where $\mathrm{TopK}$ selects tokens with the largest scores to ensure deterministic execution. Only tokens indexed by $\mathcal{I}_\ell$ are propagated to the next block. This decouples budget allocation from token selection: sensitivity profiling determines the layer-wise budgets $K_\ell$, while the composite score ranks tokens under the deployed low-bit operators to select the retained subset. The selected indices are then applied with consistent alignment across token representations, positional indices, attention masks, and the KV cache, so that keys and values are materialized and cached only for retained tokens, preserving stable autoregressive decoding under KV cache quantization.
\section{Experiments}
\label{sec:exp}
\subsection{Calibration and Benchmarks}
\label{sec:exp:data}
Post-training calibration uses a small image--text set $\mathcal{D}_{\mathrm{cal}}$ of 128 samples drawn from the LLaVA training data. We evaluate all methods under the low-bit deployment regime on standard VLM benchmarks: GQA~\cite{hudson2019gqa}, MMBench (MMB)~\cite{liu2024mmbench}, MME~\cite{fu2023mme}, POPE~\cite{li2023evaluating}, SEED-Bench~\cite{li2023seed}, and SQA$^{I}$~\cite{lu2022learn}. We report accuracy and robustness together with deployment-oriented efficiency metrics, following the official protocols and standard implementations for each benchmark.

\subsection{Implementation Details}
\label{sec:exp:impl}
All experiments are implemented in PyTorch with HuggingFace Transformers. We evaluate LLaVA-1.5-7B-hf~\cite{liu2024improved} using its default vision encoder and decoding settings unless otherwise specified. Our target regime is end-to-end W4A4 inference with a 4-bit quantized KV cache. Quantization parameters are obtained via post-training calibration on $\mathcal{D}_{\mathrm{cal}}$ using FlatQuant~\cite{sun2025flatquant}.  Since FlatQuant was developed for LLM quantization, we adapt it to the multimodal setting and apply it consistently to the vision tower, the projector, and the language model, enabling end-to-end W4A4 inference within a unified training-free implementation.

Our framework computes token-importance signals under the low-bit operators. Concretely, all metrics and pruning decisions are obtained from the W4A4 forward pass with a quantized KV cache, avoiding mismatches from full-precision scoring. Token pruning is applied only at the candidate layers $\mathcal{L}_c$. QUOTA derives the layer-wise keep ratios $\{r_\ell\}$ by profiling quantization sensitivity on $\mathcal{D}_{\mathrm{cal}}$, applying percentile-based normalization, and allocating a global drop budget with a temperature-controlled softmax. We further stabilize the allocation with a keep-ratio floor and clipping to improve robustness with small calibration sets.

\subsection{Ablation}
\label{sec:exp:ablation}
\subsubsection{Quantization Module Ablation and Real Runtime Validation.}
We first establish a feasible target deployment regime for the collaborative pipeline. We operate under end-to-end W4A4 with a quantized KV cache, and perform both token scoring and pruning under this calibrated execution form. Because the executed operators depend on which VLM components are quantized, we conduct a module-wise ablation over the vision tower, projector, and language model to identify a stable end-to-end configuration.

Many prior token pruning methods for VLMs kept the vision tower in full precision and applied pruning to the visual-token prefix inside the language model~\cite{chen2024image,yang2025topv,alvar2025divprune,zhang2025beyond,wen2025token}. This reduces cross-modal attention cost and downstream KV cache growth, and enables token-importance estimation to leverage text-conditioned interactions within the language model. To enable end-to-end low-bit deployment, we ablate which components to quantize under W4A4 with a quantized KV cache.

\begin{table}[!t]
\centering
\small
\caption{Module-wise quantization ablation on LLaVA-1.5-7B-hf under the target deployment regime. Modules specifies which modules are quantized.}
\setlength{\tabcolsep}{3.8pt}
\renewcommand{\arraystretch}{1.3}
\resizebox{.95\linewidth}{!}{%
\begin{tabular}{c| c c c c c c| c}
\toprule
\textbf{Modules} & \textbf{MMB} & \textbf{MME} & \textbf{GQA} & \textbf{POPE} & \textbf{SEED} & \textbf{SQA$^{I}$} & \textbf{Avg.} \\
\specialrule{0.1em}{2pt}{2pt}
\multirow{2}{*}{None}
& 62.37  & 1781.12 & 60.45  & 85.89  & 64.01  & 66.09  & \multirow{2}{*}{100\%} \\
& 100\% & 100\%  & 100\% & 100\% & 100\% & 100\% & \\
\midrule
\midrule
\multirow{2}{*}{LLM}
& 56.70  & 1700.75 & 59.10  & 85.37  & 61.72  & 64.50  & \multirow{2}{*}{\second{96.26\%}} \\
& \best{90.91\%} & 95.49\%  & 97.77\% & \best{99.39\%} & 96.42\% & \best{97.59\%} & \\
\midrule
\multirow{2}{*}{VT\&LLM}
& 56.19  & 1702.59 & 59.25  & 84.61  & 62.50  & 62.57  & \multirow{2}{*}{95.75\%} \\
& \second{90.09\%} & \second{95.59\%}  & \best{98.01\%} & 98.51\% & \second{97.64\%} & 94.67\% & \\
\midrule
\multirow{2}{*}{VT\&PJ\&LLM}
& 56.10  & 1749.30 & 59.25  & 84.98  & 62.73 & 63.16  & \multirow{2}{*}{\best{96.45\%}} \\
& 89.95\% & \best{98.21\%}  & \best{98.01\%} & \second{98.94\%} & \best{98.00\%} & \second{95.57\%} & \\
\bottomrule
\end{tabular}
}
\label{tab:quant_ablation_modules}
\end{table}
Table~\ref{tab:quant_ablation_modules} summarizes the module-wise ablation used to select an end-to-end low-bit operating regime. Quantizing the language model alone retained 96.26\% average performance relative to the full-precision baseline. Quantizing the vision tower in addition led to a larger drop, most noticeably on SQA$^{I}$, which is consistent with higher sensitivity in vision-intensive reasoning. When the projector was also quantized, average retention slightly improved to 96.45\%, suggesting that calibrating the cross-modal interface jointly with the language model can yield more consistent low-bit behavior. Based on these results, we use end-to-end W4A4 quantization for the vision tower, projector, and language model, with a quantized KV cache, as the default regime. All token pruning experiments, stage-wise baselines, and the proposed collaborative pipeline use this same calibrated low-bit execution form.

Beyond the accuracy ablation above, we further validate the practical deployability of the selected low-bit quantized path by implementing a real quantized inference runtime with custom CUDA kernels. Under true INT4 execution, the on-disk model size is reduced from 14.13\,GB to 3.97\,GB on LLaVA-1.5-7B and from 16.6\,GB to 5.86\,GB on Qwen2.5-VL-7B, accompanied by clear reductions in peak GPU memory and average GPU power. The real-quantized results also remain broadly consistent with the fake-quant evaluation, indicating that the selected low-bit regime transfers reliably from calibration-time analysis to actual deployment. Detailed per-task runtime measurements are provided in the supplementary material.

\subsubsection{Pruning Ratio Ablation.}
We study the effect of the final keep ratio under a fixed low-bit regime. We fix W4A4 with a quantized KV cache, the candidate set $\mathcal{L}_c$, and the scoring metrics, and vary the final keep ratio by changing QUOTA's keep-ratio floor $p_{\min}$, while keeping all other settings unchanged.
\begin{table}[!t]
\centering
\small
\caption{Pruning ratio ablation under the same low-bit operating regime. The second line reports retention relative to the full precision model without pruning.}
\setlength{\tabcolsep}{3.8pt}
\renewcommand{\arraystretch}{1.3}
\resizebox{.95\linewidth}{!}{%
\begin{tabular}{c | c c c c c c | c}
\toprule
\textbf{Final keep ratio} & \textbf{MMB} & \textbf{MME} & \textbf{GQA} & \textbf{POPE} & \textbf{SEED} & \textbf{SQA$^{I}$} & \textbf{Avg.} \\
\midrule
\multirow{2}{*}{50\%}
& 57.59 & 1667.80 & 58.13 & 85.12 & 62.53 & 63.86 & \multirow{2}{*}{95.93\%} \\
& 92.34\% & 93.64\% & 96.16\% & 99.10\% & 97.69\% & 96.63\% & \\
\midrule
\multirow{2}{*}{40\%}
& 57.54 & 1689.33 & 58.14 & 84.54 & 62.31 & 63.31 & \multirow{2}{*}{95.81\%} \\
& 92.26\% & 94.85\% & 96.18\% & 98.43\% & 97.34\% & 95.79\% & \\
\midrule
\multirow{2}{*}{30\%}
& 57.47 & 1688.72 & 57.97 & 84.83 & 62.10 & 62.96 & \multirow{2}{*}{95.65\%} \\
& 92.14\% & 94.81\% & 95.90\% & 98.77\% & 97.02\% & 95.26\% & \\
\midrule
\multirow{2}{*}{20\%}
& 53.87 & 1617.66 & 56.54 & 83.79 & 60.87 & 63.61 & \multirow{2}{*}{93.27\%} \\
& 86.37\% & 90.82\% & 93.53\% & 97.56\% & 95.09\% & 96.25\% & \\
\midrule
\multirow{2}{*}{10\%}
& 53.01 & 1608.78 & 54.62 & 82.46 & 59.54 & 62.52 & \multirow{2}{*}{91.55\%} \\
& 84.99\% & 90.32\% & 90.36\% & 96.01\% & 93.02\% & 94.60\% & \\
\midrule
\multirow{2}{*}{5\%}
& 47.51 & 1481.74 & 52.07 & 80.34 & 56.64 & 61.87 & \multirow{2}{*}{86.86\%} \\
& 76.17\% & 83.19\% & 86.14\% & 93.54\% & 88.49\% & 93.61\% & \\
\bottomrule
\end{tabular}
}
\label{tab:prune_ratio_ablation}
\end{table}

Table~\ref{tab:prune_ratio_ablation} shows that performance is stable at moderate pruning levels under the calibrated low-bit regime. Retaining 30\% of visual tokens achieves 95.65\% average retention relative to the full-precision baseline, and the average retention changes by only 0.28 between 50\% and 30\%, indicating substantial redundancy in the visual-token prefix under low-bit inference. A clear degradation occurs below a 30\% final keep ratio: average retention drops from 95.65\% (30\%) to 93.27\% (20\%) and 91.55\% (10\%). POPE remains relatively stable, while MMBench, MME, and GQA degrade more noticeably, indicating a stronger reliance on fine-grained visual context. At 5\%, retention further decreases to 86.86\%, suggesting an information-bottleneck regime where performance becomes highly sensitive to additional token removal. Based on this ablation, we use a 30\% final keep ratio as the default setting in the main comparisons. This choice provides substantial token reduction while maintaining strong accuracy retention under low-bit inference.

\subsection{Comparison: Stage-wise Baselines and Collaborative Pipeline}
We compare the proposed collaborative pipeline with representative stage-wise baselines to characterize how token pruning interacts with low-bit quantization under the same deployment regime. All pruning variants use the candidate layer set $\mathcal{L}_c$ in Sec.~\ref{subsec:budgets} and apply a stage-wise keep-ratio schedule that ends with a 30\% final keep ratio at the last candidate layer. Token importance is estimated from the magnitude cue and the intra-visual/inter-modal attention cues described in Sec.~\ref{subsec:score}. For quantization, we use the end-to-end W4A4 configuration selected by the module-wise ablation in Table~\ref{tab:quant_ablation_modules}, which quantizes the vision tower, projector, and language model and uses a quantized KV cache.

Quantize-then-Prune calibrates low-bit quantization on the unpruned graph and applies the pruning recipe only at deployment. Prune-then-Quant applies the same recipe first to define the pruned execution form, and then calibrates quantization on the pruned graph using activations from the retained-token subgraph. In contrast, our collaborative pipeline derives the layer-wise keep-ratio schedule from low-bit calibration signals via QUOTA and evaluates token importance under the deployed quantized operators with a quantized KV cache. This aligns pruning decisions with the calibrated low-bit execution form.

\begin{table}[!t]
\centering
\small
\caption{Stage-wise baselines and the proposed collaborative pipeline under the default low-bit operating regime. The first three rows provide reference baselines (full precision, quant-only, and prune-only). The second row of each setting reports retention relative to the full-precision model without token pruning.}
\setlength{\tabcolsep}{3.8pt}
\renewcommand{\arraystretch}{1.3}
\resizebox{.95\linewidth}{!}{%
\begin{tabular}{c | c c c c c c | c}
\toprule
\textbf{Variant} & \textbf{MMB} & \textbf{MME} & \textbf{GQA} & \textbf{POPE} & \textbf{SEED} & \textbf{SQA$^{I}$} & \textbf{Avg.} \\
\midrule
\multirow{2}{*}{LLaVA-1.5-7B-hf}
& 62.37  & 1781.12 & 60.45  & 85.89  & 64.01 & 66.09  & \multirow{2}{*}{100\%} \\
& 100\% & 100\%  & 100\% & 100\% & 100\% & 100\% & \\
\midrule
\multirow{2}{*}{Quant-only}
& 56.10  & 1749.30 & 59.25  & 84.98  & 62.73 & 63.16  & \multirow{2}{*}{96.45\%} \\
& 89.95\% & 98.21\%  & 98.01\% & 98.94\% & 98.00\% & 95.57\% & \\
\midrule
\multirow{2}{*}{Prune-only}
&  60.82 & 1752.15 &  58.46 & 85.10  & 62.13 &  64.85 & \multirow{2}{*}{97.81\%} \\
& 97.51\% &  98.37\% & 96.71\% & 99.08\% & 97.06\% & 98.12\% & \\
\midrule
\midrule
\multirow{2}{*}{Quant then Prune}
& 56.70  & 1652.74 & 56.82  & 84.28  & 61.35 &  62.22 & \multirow{2}{*}{94.30\%} \\
& \second{90.91\%}  & 92.79\%  & 94.00\% & 98.13\% & \second{95.84\%} & 94.14\% & \\
\midrule
\multirow{2}{*}{Prune then Quant}
& 56.44 & 1654.98 & 56.86  & 84.54  & 60.50 &  63.21 & \multirow{2}{*}{\second{94.34\%}} \\
& 90.49\% & \second{92.92\%} & \second{94.06\%} & \second{98.43\%} & 94.52\% &  \best{95.64\%} & \\
\midrule
\multirow{2}{*}{Ours Collaborative}
& 57.47 & 1688.72 & 57.97 & 84.83 & 62.10 & 62.96 & \multirow{2}{*}{\best{95.65\%}} \\
& \best{92.14\%} &  \best{94.81\%} & \best{95.90\%} & \best{98.77\%} & \best{97.02\%} & \second{95.26\%}  & \\
\bottomrule
\end{tabular}
}
\label{tab:stagewise_compact}
\end{table}
Table~\ref{tab:stagewise_compact} reports accuracy under the same end-to-end W4A4 operating regime. Quantization-only retained 96.45\% of the full-precision baseline on average, while pruning-only with a 30\% final keep ratio remained relatively stable at 97.81\%, indicating that each component in isolation introduced only a moderate degradation at the target compression level. In contrast, the stage-wise combinations exhibited a clear coupling effect under the same operating regime: both Quantize-then-Prune and Prune-then-Quant dropped to about 94.3\% average retention, substantially worse than either quantization-only or pruning-only. This non-additivity suggests that token pruning and low-bit quantization interact through a calibration--execution mismatch. Calibrating quantization on the unpruned graph and pruning afterward makes the calibrated low-bit parameters less transferable to the pruned execution form, while calibrating on the pruned graph alone does not fully restore robustness when token importance is evaluated under aggressive low-bit operators.

The proposed collaborative pipeline outperforms both stage-wise baselines on average and achieves the highest average retention of 95.65\%. It attains the best performance on MMBench, MME, GQA, POPE, and SEED, and remains competitive on SQA$^{I}$, where Prune-then-Quant is slightly higher. These results support our central claim that token pruning should be aligned with the calibrated low-bit operating regime. By deriving the layer-wise keep-ratio schedule from low-bit calibration signals and evaluating token importance under the same quantized operators used, the collaborative design reduces the brittleness of naive stage-wise integration while maintaining the target pruning ratio.

\subsection{Efficiency and Accuracy Retention}
\label{sec:exp:eff}
Building on the coupling analysis in Table~\ref{tab:stagewise_compact}, Table~\ref{tab:efficiency} relates robustness under low-bit inference to token reduction and compute savings. We report token counts and estimate GFLOPs ratios using the text token length observed on POPE as a proxy for decoding workload. Because the cost of cross-modal attention and autoregressive decoding grows with the text sequence length~\cite{huang2024ivtp}, and many prior works did not disclose generation lengths or decoding configurations, the resulting GFLOPs ratios should be viewed as approximate and are intended to capture overall trends rather than exact cross-paper comparisons.

Token-pruning baselines reduce compute primarily by shortening the visual-token prefix, which directly lowers the reported GFLOPs ratio under our measurement protocol. Table~\ref{tab:efficiency} shows that more aggressive token reduction generally yields larger compute savings. DivPrune reduces GFLOPs to 21.73\% with 128 visual tokens while maintaining strong FP16 accuracy on average. However, token count alone does not fully determine the GFLOPs ratio, since the realized savings also depend on the depth at which tokens are removed and the extent to which subsequent cross-modal attention is avoided. In contrast, quantization-only baselines keep the token sequence unchanged and therefore show no reduction in the reported GFLOPs ratio. Their benefit lies in reduced parameter precision and lower memory traffic under low-bit arithmetic, which can improve deployment efficiency even when the FLOPs count is unchanged, but may come with accuracy degradation depending on the quantizer and bitwidth. 

\begin{table}[!t]
\centering
\small
\setlength{\tabcolsep}{3.2pt}
\renewcommand{\arraystretch}{1.3}
\caption{Accuracy retention and computational cost for token pruning, quantization, and the joint setting.}
\resizebox{\linewidth}{!}{%
\begin{tabular}{l c c c c c c c c c}
\toprule
\textbf{Method} & \textbf{Pre.} & \textbf{Token} & \textbf{GFLOPs} & \textbf{MMB} & \textbf{MME} & \textbf{GQA} & \textbf{POPE} & \textbf{SQA$^{I}$} & \textbf{Avg} \\
\midrule
Original & FP16 & 576 & 100.00\% & 100.00\% & 100.00\% & 100.00\% & 100.00\% & 100.00\% & 100.00\% \\
\midrule
\multicolumn{10}{l}{\textbf{Token Pruning}} \\
\midrule
VTW~\cite{lin2025boosting} & FP16 & 236 & 47.32\% & 98.60\% & 97.64\% & 82.74\% & 95.69\% & 99.14\% & 93.80\% \\
PDrop~\cite{xing2024pyramiddrop}    & FP16 & 192 & 34.03\% & 95.80\% & 92.60\% & 92.10\% & 97.44\% & 98.56\% & 95.30\% \\
FastV~\cite{chen2024image}    & FP16 & 172 & 43.20\% & 95.80\% & 96.97\% & 92.90\% & 94.41\% & 99.28\% & 95.87\% \\
DivPrune~\cite{alvar2025divprune} & FP16 & 128 & 21.73\% & 96.58\% & 93.03\% & 94.84\% & 99.18\% & 98.56\% & 96.44\% \\
\midrule
\multicolumn{10}{l}{\textbf{Quantization only}} \\
\midrule
GPTQ~\cite{frantar2023gptq}  & W4A16 & 576 & 100.00\% & 91.80\% & 96.61\% & 99.03\% & 91.35\% & -- & 94.59\% \\
AWQ~\cite{lin2024awq}   & W4A16 & 576 & 100.00\% & 94.32\% & 97.04\% & 99.19\% & 92.19\% & -- & 95.69\% \\
GRACE~\cite{chen2026gated} & W4A16 & 576 & 100.00\% & 101.09\% & -- & 98.71\% & 99.19\% & 97.91\% & 99.23\% \\
LUQ~\cite{bhatnagar2025luq}   & W2.54A16 & 576 & 100.00\% & 73.66\% & 87.20\% & 93.87\% & 89.54\% & -- & 86.07\% \\
\midrule
\rowcolor{black!8}
\textbf{Ours} & {W4A4KV4} & {173} & {40.89\%} & {92.14\%} & {94.81\%} & {95.90\%} & {98.77\%} & {95.26\%} & {95.38\%} \\
\bottomrule
\end{tabular}
}
\label{tab:efficiency}
\end{table}

Under the joint setting, our pipeline combines both sources of efficiency: it retains 173 visual tokens and reduces the estimated GFLOPs to 40.89\% while maintaining 95.38\% average retention under W4A4KV4. This operating point lies in a similar compute range to representative FP16 token pruning baselines (e.g., FastV), yet remains robust under aggressive low-bit inference. Together with Table~\ref{tab:stagewise_compact}, these results suggest that aligning budget derivation and token selection with the calibrated low-bit operators enables effective token reduction without the degradation observed in stage-wise combinations.

\subsubsection{Qualitative Visualization.}
\begin{figure*}[t]
  \centering
  \includegraphics[width=\textwidth]{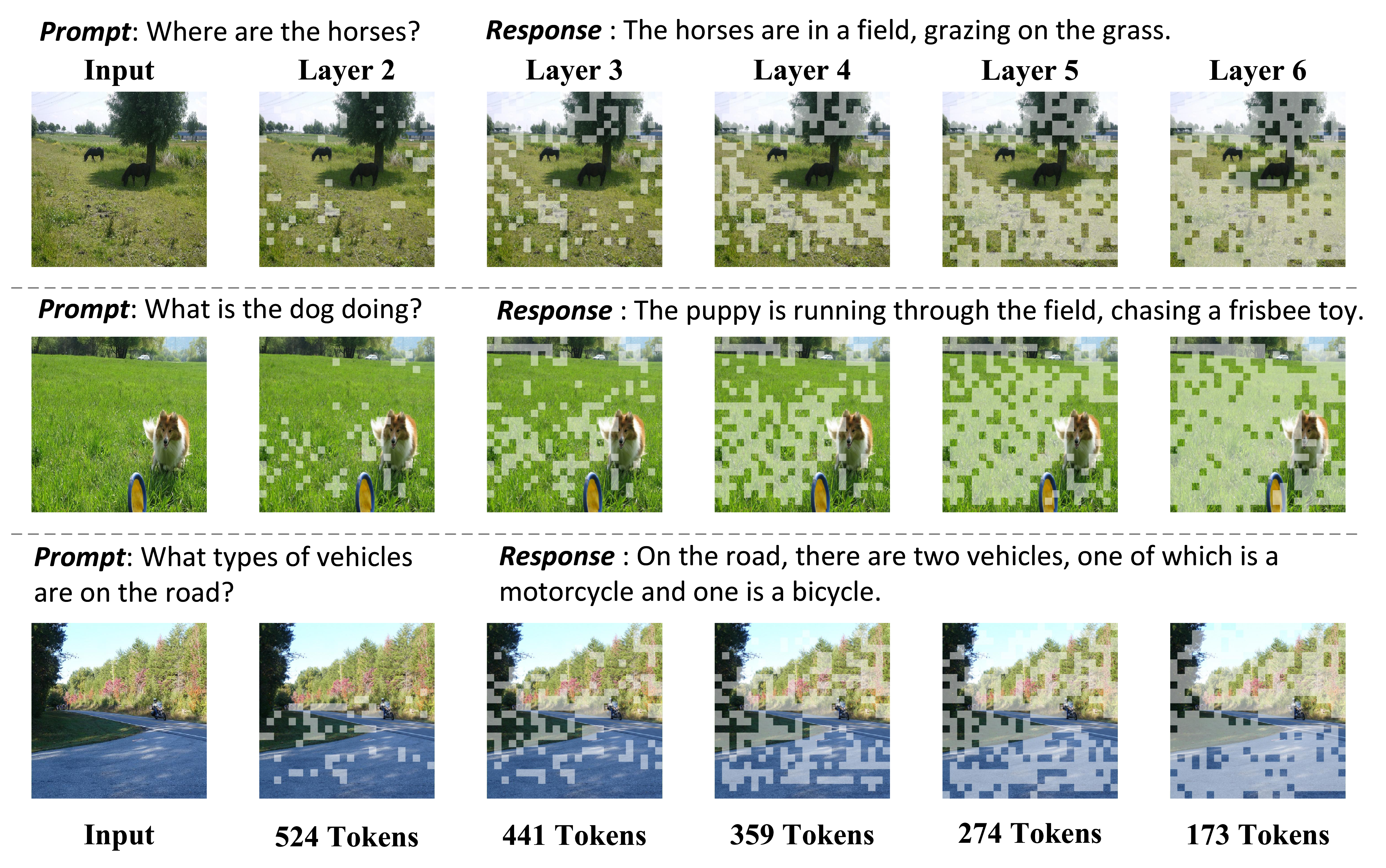}
  \caption{Qualitative visualization of visual-token pruning under low-bit inference with a quantized KV cache. Retained visual tokens are overlaid across candidate layers, with token counts reported for each layer.}
  \label{fig:pruning_vis}
\end{figure*}

Figure~\ref{fig:pruning_vis} visualizes how visual-token pruning behaves under the same low-bit operating regime. As dictated by the layer-wise keep-ratio schedule, the number of retained visual tokens decreases across candidate layers, while the retained tokens progressively concentrate on semantically relevant regions. This concentration suggests that the budgeted selection preserves salient visual evidence needed for cross-modal reasoning.

Across diverse prompts, the model produces coherent responses despite substantial reduction of the visual-token prefix. This qualitative behavior is consistent with QUOTA-derived budgeting and quantization-consistent token selection retaining informative tokens under deployed low-bit operators.

\section{Conclusion}
\label{sec:conclusion}
Recent VLMs are powerful but costly to deploy, as inference is often dominated by the long visual-token prefix and the growing KV cache during decoding. Low-bit quantization and visual-token pruning are complementary for reducing memory traffic and sequence length, yet their interaction under aggressive low-bit execution remains underexplored.
To this end, we study joint low-bit quantization and visual-token pruning for deployable VLM inference, and show that naive stage-wise integration suffered from a calibration--execution mismatch under aggressive low-bit settings. To address this issue, we propose a collaborative pipeline that aligns token pruning with the calibrated low-bit operating regime. The pipeline introduces QUOTA to convert low-bit calibration signals into a deterministic layer-wise pruning recipe, and performs budgeted token selection under the deployed quantized operators with a quantized KV cache. 

Across multiple VLM benchmarks, the proposed pipeline improves robustness over representative stage-wise baselines under the same W4A4 regime. Ablations further show a stable operating region at moderate pruning levels, where retaining only 30\% of visual tokens maintains strong accuracy retention. Future work will explore more principled co-design of pruning and quantization for low-bit VLM deployment, including adaptive budgeting and selection strategies, broader low-bit settings, and deployment-oriented measurements beyond the current calibration-driven recipe and deterministic selection pipeline.

\section*{Acknowledgements}
This work is supported in part by the National Natural Science Foundation of China (No. 62576176), in part by the Tianjin Science and Technology Major Project (No. 25ZXRGGX00120), and in part by the Fundamental Research Funds for the Central Universities (Nankai University, No. 070-63253235). The computational resources are supported by the Supercomputing Center of Nankai University (NKSC).


%
%

%% file: supp2all.tex
\section{Overview}
This supplementary document provides additional details and discussion to complement the main manuscript, \emph{Towards Joint Quantization and Token Pruning of Vision-Language Models}. The main paper introduced a collaborative deployment pipeline that aligns visual-token pruning with the calibrated low-bit operating regime, motivated by the observation that naive stage-wise combinations of quantization and pruning can suffer from a calibration--execution mismatch under aggressive low-bit inference. In particular, the proposed framework uses QUOTA to convert low-bit calibration signals into a deterministic layer-wise pruning recipe and performs token selection under deployed W4A4 operators with a quantized KV cache. This supplementary document expands several aspects that were condensed in the main paper due to space limitations.

The supplementary material is organized to further clarify the main components of the framework and to make the overall pipeline easier to reproduce and interpret. We first revisit the unified low-bit inference pipeline and describe how calibration, recipe construction, and deployment-time pruning interact. We then provide additional explanation of the candidate-layer selection strategy, the sensitivity-conditioned token-budget allocation, and the composite token-selection score used under low-bit execution. Next, we summarize the experimental protocol and clarify the settings used in the main comparisons, including the low-bit deployment regime, the benchmark suite, and the baseline definitions.

We also provide further discussion of the main empirical findings reported in the paper. These include the module-wise quantization ablation used to determine the end-to-end W4A4 configuration, the pruning-ratio ablation that motivated the default 30\% final keep ratio, and the comparison between the proposed collaborative pipeline and representative stage-wise baselines. Finally, we include additional remarks on the reported efficiency trends and qualitative visualizations, with a focus on how the unified design preserves robustness under substantial visual-token reduction.

Taken together, this supplementary document provides a more detailed view of the design choices and experimental observations behind the main paper, while reinforcing the central conclusion that pruning decisions should remain consistent with the calibrated low-bit operating regime.

\section{Additional Implementation Details for Qwen2.5-VL}
\label{supp:impl}
\subsection{Calibration Set}
The calibration set should remain statistically close to the input distribution encountered during deployment. We therefore randomly sample 128 instances from the original LLaVA training set. Using samples drawn directly from the training corpus helps preserve consistency between the calibration distribution and the data distribution on which the model originally learned its multimodal representations. For reproducibility, the sampled subset is stored as a static JSON file, ensuring that exactly the same calibration samples are used across different runs and across different quantization methods. Each instance includes both an image and a text instruction, allowing the calibration process to better approximate realistic multimodal interaction scenarios.

While the main paper focuses on aggressive low-bit settings, particularly INT4 quantization with a quantized KV cache, the proposed framework is not restricted to a single precision level. More generally, the calibration-driven pipeline can be instantiated under different low-bit regimes. In this work, however, we instantiate it under INT4, with calibration and deployment conducted under the same target quantized operators. The framework therefore provides a unified recipe-generation and token-pruning mechanism for different low-bit deployment regimes.

\subsection{Details of GFLOPs Estimation}
We report GFLOPs as a deployment-oriented efficiency proxy to quantify how visual-token pruning changes the prefill-side computational cost of the VLM. In our setting, this stage accounts for a substantial portion of the LLM-side computation, since the long visual-token prefix is processed before autoregressive decoding begins. Our estimate is designed to capture the computation reduction caused by shortening the executed visual-token prefix in the multimodal LLM, rather than to provide an exact end-to-end accounting of autoregressive decoding cost.

Let $V_0$ denote the reference number of visual tokens after the projector, and let $\hat{V}_{\ell}$ denote the executed number of visual tokens at LLM layer $\ell$ under the deployed pruning recipe. Before the first pruning layer, we have $\hat{V}_{\ell}=V_0$. Once pruning is applied at a candidate layer, only the retained tokens are propagated to subsequent layers, so later layers operate on a shorter visual-token prefix. Therefore, the computational reduction is cumulative across depth and depends not only on the final retained token count, but also on where pruning is applied.

We estimate the prefill-oriented VLM computation as
\begin{equation}
\mathcal{F}_{\mathrm{VLM}}
=
\mathcal{F}_{\mathrm{VT}}
+
\mathcal{F}_{\mathrm{PJ}}
+
\sum_{\ell=1}^{L}
\mathcal{C}_{\ell}(\hat{V}_{\ell}, T),
\end{equation}

where $\mathcal{F}_{\mathrm{VT}}$ and $\mathcal{F}_{\mathrm{PJ}}$ denote the FLOPs of the vision tower and projector, respectively, $L$ is the number of LLM layers, and $T$ is a fixed text-length proxy used in the evaluation protocol. Since pruning is applied after the projector in the multimodal LLM, $\mathcal{F}_{\mathrm{VT}}$ and $\mathcal{F}_{\mathrm{PJ}}$ remain unchanged across pruning settings, and the GFLOPs reduction mainly comes from the change in $\mathcal{C}_{\ell}(\hat{V}_{\ell}, T)$ in the subsequent LLM layers.

For the unpruned baseline, the executed visual-token length is fixed to $V_0$ for all layers, yielding
\begin{equation}
\mathcal{F}_{\mathrm{base}}
=
\mathcal{F}_{\mathrm{VT}}
+
\mathcal{F}_{\mathrm{PJ}}
+
\sum_{\ell=1}^{L}
\mathcal{C}_{\ell}(V_0, T).
\end{equation}
We then report the GFLOPs ratio as
\begin{equation}
\mathrm{GFLOPs\ Ratio}
=
\frac{\mathcal{F}_{\mathrm{VLM}}}{\mathcal{F}_{\mathrm{base}}}
\times 100\%.
\end{equation}

In practice, $\mathcal{C}_{\ell}(\hat{V}_{\ell}, T)$ is computed from the executed sequence length at each layer. If we denote the layer-wise total sequence length by
\begin{equation}
S_{\ell} = T + \hat{V}_{\ell},
\end{equation}
then the dominant transformer cost at layer $\ell$ can be viewed as a function of $S_{\ell}$, including both attention-related computation and token-wise projection/MLP computation. Under this protocol, reducing $\hat{V}_{\ell}$ lowers the computation of the current layer and all later layers that operate on the shortened sequence.

For decoding-related comparisons, we use the text token length observed on POPE as a fixed workload proxy, following the protocol in the main paper. This design allows different methods to be compared under a consistent text-generation assumption. As a result, the reported GFLOPs should be interpreted primarily as a prefill-oriented computational indicator rather than an exact end-to-end runtime measurement. In particular, actual latency also depends on memory traffic, kernel efficiency, hardware behavior, and low-bit implementation details, which are not fully captured by FLOPs alone.

This estimation protocol also explains why token count alone does not fully determine the reported GFLOPs ratio. Even if two methods retain a similar final number of visual tokens, their GFLOPs can still differ if pruning happens at different depths or avoids different amounts of downstream cross-modal computation. Earlier pruning typically produces larger cumulative savings because the reduced visual-token prefix is inherited by more subsequent layers.


\section{Additional Ablation Studies}
\label{supp:ablations}
\subsection{Ablation on Metric Weights}
Our token-selection score combines four complementary signals, including the inter-modal attention term, the activation-magnitude term, the intra-visual attention term, and a quantization-related term. The default metric weights are chosen to reflect the expected role of each signal in VLM token selection. We assign the largest weight to the inter-modal attention term, since VLM inference fundamentally depends on cross-modal interaction between text queries and visual tokens. A visual token that receives stronger inter-modal attention is more likely to contribute directly to downstream language reasoning and response generation. The activation-magnitude term is assigned the second largest weight because token representations with larger magnitudes tend to carry stronger responses and have a greater influence on subsequent transformations, making it a useful salience cue.

We further include a quantization-related term to explicitly account for low-bit sensitivity. Intuitively, a larger discrepancy between a token representation and its quantized counterpart indicates that this token is harder to represent stably under low-bit quantization, and is therefore more vulnerable to distortion introduced by the deployed quantized operators. In contrast, tokens that are more easily preserved under quantization are less likely to suffer critical information loss during low-bit execution. This signal is not treated as a standalone measure of semantic importance. Instead, under a limited pruning budget, it serves as a robustness-aware auxiliary cue that helps avoid discarding tokens whose information may be especially fragile in the target low-bit regime. Finally, the intra-visual attention term is assigned a smaller but non-zero weight because interactions among visual tokens still provide useful structural information.

To study the effect of the metric weights, we conduct a targeted sensitivity analysis rather than an exhaustive search over all possible combinations. Since the four coefficients are defined on a normalized simplex, a full search would be prohibitively expensive. Our goal here is not to identify a globally optimal coefficient set, but to evaluate whether each metric contributes to the final token-selection score and whether the weighting scheme used in the main paper is reasonably robust.

Specifically, we consider two types of perturbations. First, we perform a leave-one-out ablation by setting one metric weight to zero at a time and redistributing the removed weight uniformly to the remaining metrics. Second, we perform a coarse two-dimensional sweep over selected configurations to examine whether the performance remains stable under moderate weight perturbations around the main setting. Based on the design considerations discussed above, we use an inter-modal-dominant weighting scheme in the main experiments, with the activation-magnitude term assigned the second largest weight, the quantization-related term given a non-negligible auxiliary weight, and the intra-visual attention term assigned a smaller but non-zero coefficient.

\begin{table}[!t]
\centering
\small
\caption{Ablation on the metric weights in the composite token-selection score. We report the default setting, leave-one-out removal of each metric, and a coarse two-dimensional sweep around the default setting. The second line of each configuration reports retention relative to the full-precision baseline. Best retention is highlighted in bold, and the second-best retention is underlined.}
\setlength{\tabcolsep}{3.8pt}
\renewcommand{\arraystretch}{1.3}
\resizebox{.95\linewidth}{!}{%
\begin{tabular}{c | c c c c | c c c | c}
\toprule
\multirow{2}{*}{\textbf{Variant}} 
& \multicolumn{4}{c|}{\textbf{Metric weights}} 
& \multirow{2}{*}{\textbf{MMB}} 
& \multirow{2}{*}{\textbf{MME}} 
& \multirow{2}{*}{\textbf{SQA$^{I}$}} 
& \multirow{2}{*}{\textbf{Avg.}} \\
& \textbf{Inter} & \textbf{Mag.} & \textbf{Intra} & \textbf{Quant.} & & & & \\
\midrule
\multirow{2}{*}{Full precision}
& \multicolumn{4}{c|}{\multirow{2}{*}{--}}
& 62.37 & 1781.12 & 66.09 & \multirow{2}{*}{100\%} \\
& \multicolumn{4}{c|}{}
& 100\% & 100\% & 100\% & \\
\midrule
\multirow{2}{*}{Default} & \multirow{2}{*}{45.00\%} & \multirow{2}{*}{25.00\%} & \multirow{2}{*}{10.00\%} & \multirow{2}{*}{20.00\%} & 57.47 & 1688.72 & 62.96 & \multirow{2}{*}{\textbf{94.07\%}} \\
 & & & & & \textbf{92.14\%} & \second{94.81\%} & 95.26\% & \\
\midrule
\multirow{2}{*}{Inter removed} & \multirow{2}{*}{0.00\%} & \multirow{2}{*}{40.00\%} & \multirow{2}{*}{25.00\%} & \multirow{2}{*}{35.00\%} & 55.84 & 1642.79 & 63.41 & \multirow{2}{*}{92.57\%} \\
 & & & & & 89.53\% & 92.23\% & 95.94\% & \\
 \midrule
\multirow{2}{*}{Mag. removed} & \multirow{2}{*}{53.33\%} & \multirow{2}{*}{0.00\%} & \multirow{2}{*}{18.33\%} & \multirow{2}{*}{28.33\%} & 57.04 & 1654.61 & 62.94 & \multirow{2}{*}{93.20\%} \\
 & & & & & 91.45\% & 92.90\% & 95.23\% & \\
 \midrule
\multirow{2}{*}{Intra removed} & \multirow{2}{*}{48.33\%} & \multirow{2}{*}{28.33\%} & \multirow{2}{*}{0.00\%} & \multirow{2}{*}{23.33\%} & 55.84 & 1641.96 & 63.21 & \multirow{2}{*}{92.45\%} \\
 & & & & & 89.53\% & 92.19\% & 95.64\% & \\
 \midrule
\multirow{2}{*}{Quant. removed} & \multirow{2}{*}{51.67\%} & \multirow{2}{*}{31.67\%} & \multirow{2}{*}{16.67\%} & \multirow{2}{*}{0.00\%} & 57.21 & 1622.62 & 63.76 & \multirow{2}{*}{93.10\%} \\
 & & & & & 91.73\% & 91.10\% & \textbf{96.47\%} & \\
\midrule
\multirow{2}{*}{Inter$\downarrow$, Quant.$\downarrow$} & \multirow{2}{*}{35.00\%} & \multirow{2}{*}{35.00\%} & \multirow{2}{*}{20.00\%} & \multirow{2}{*}{10.00\%} & 57.39 & 1671.27 & 63.11 & \multirow{2}{*}{93.78\%} \\
 & & & & & \second{92.02\%} & 93.83\% & 95.49\% & \\
 \midrule
\multirow{2}{*}{Inter$\downarrow$, Quant.$\uparrow$} & \multirow{2}{*}{35.00\%} & \multirow{2}{*}{25.00\%} & \multirow{2}{*}{10.00\%} & \multirow{2}{*}{30.00\%} & 56.87 & 1682.47 & 63.66 & \multirow{2}{*}{\second{93.99\%}} \\
 & & & & & 91.18\% & 94.46\% & \second{96.32\%} & \\
 \midrule
\multirow{2}{*}{Inter$\uparrow$, Quant.$\downarrow$} & \multirow{2}{*}{55.00\%} & \multirow{2}{*}{25.00\%} & \multirow{2}{*}{10.00\%} & \multirow{2}{*}{10.00\%} & 56.19 & 1656.24 & 62.57 & \multirow{2}{*}{92.58\%} \\
 & & & & & 90.09\% & 92.99\% & 94.67\% & \\
 \midrule
\multirow{2}{*}{Inter$\uparrow$, Quant.$\uparrow$} & \multirow{2}{*}{55.00\%} & \multirow{2}{*}{15.00\%} & \multirow{2}{*}{0.00\%} & \multirow{2}{*}{30.00\%} & 56.44 & 1688.99 & 62.72 & \multirow{2}{*}{93.41\%} \\
 & & & & & 90.49\% & \textbf{94.83\%} & 94.90\% & \\ 
\bottomrule
\end{tabular}
}
\label{tab:weight_ablation}
\end{table}

Table~\ref{tab:weight_ablation} reports the effect of different weight settings on the composite token-selection score. The default configuration achieves the best average retention, reaching 94.07\%, which supports the weighting scheme adopted in the main experiments. More importantly, the coarse two-dimensional sweep reveals a favorable high-performing basin around the default setting: several nearby configurations remain competitive, but none surpasses the default configuration on average. This suggests that the chosen weighting scheme is not an isolated heuristic point, but lies in a relatively stable local region of the tested weight grid. 

The leave-one-out ablation further shows that all four metrics contribute to the final score. Removing any single term leads to a drop in average retention relative to the default configuration. In particular, removing the inter-modal attention term reduces the average retention to 92.57\%, while removing the intra-visual attention term leads to the lowest average retention of 92.45\%. These results indicate that both cross-modal interaction and visual-token structural context play important roles in identifying informative tokens under the low-bit regime. Removing the activation-magnitude term also degrades the overall performance to 93.20\%, confirming that token salience remains a useful complementary cue. Removing the quantization-related term yields 93.10\% average retention. Although this variant achieves the best retention on SQA$^{I}$, it performs worst on MME, suggesting that the quantization-related term mainly acts as a robustness-aware auxiliary signal rather than a universally dominant semantic cue.

Within the tested coarse grid, the setting with reduced inter-modal attention weight and increased quantization-related weight achieves the second-best average retention of 93.99\%, which is very close to the default result. Together with the other nearby configurations, this trend suggests that the tested grid contains a relatively smooth performance basin rather than a highly fragile single-point optimum. At the same time, increasing the inter-modal attention weight to 55\% leads to a clear drop in average retention, regardless of whether the quantization-related term is reduced or increased. This indicates that, although the inter-modal attention term should remain the dominant component, over-emphasizing it can suppress other useful cues and move the weighting scheme away from the favorable basin.
\section{Additional Experimental Results}
\label{supp:extra_results}
\subsection{Experimental Setup}
We further evaluated the proposed collaborative pipeline on Qwen2.5-VL-7B~\cite{bai2025qwen25vl} to examine whether the main findings of this work generalize to a different VLM backbone. Unless otherwise specified, we followed the same experimental protocol as in the main paper. Specifically, we retained the same low-bit evaluation principle and deployed Qwen2.5-VL-7B under a W4A4 low-bit setting, with the remaining deployment configuration chosen according to the stable implementation used for this backbone. As in the main setting, visual-token pruning was applied in the LLM component of the model.

To ensure a fair comparison, we evaluated on the same six benchmarks used in the main paper, namely GQA~\cite{hudson2019gqa}, MMBench~\cite{liu2024mmbench}, MME~\cite{fu2023mme}, POPE~\cite{li2023evaluating}, SEED-Bench~\cite{li2023seed}, and SQA$^{I}$~\cite{lu2022learn}. All benchmark evaluations were conducted using the LMMs-Eval~\cite{zhang2025lmms} toolkit under the same benchmark suite and evaluation protocol as in the main experiments. Except for backbone-specific adaptations required by Qwen2.5-VL-7B, the remaining settings were kept as consistent as possible with those of the main paper, including the same minimum final keep ratio of 30\% and the same calibration-driven recipe construction principle. In the Qwen2.5-VL-7B experiments, the layer-wise pruning plan was derived directly during calibration.

\subsection{Implementation Details}
Visual-token pruning was applied in the LLM component of the model, following the same overall collaborative pipeline as in the main paper. Backbone-specific modifications were limited to the parts required to support Qwen2.5-VL-7B within the same calibration-driven pruning and low-bit deployment framework.

\subsubsection{Calibration Data}
Unlike LLaVA, Qwen2.5-VL did not release its full training corpus for direct reuse in calibration. According to the technical report, Qwen2.5-VL was trained on a substantially expanded multimodal corpus at the trillion-token scale, constructed from cleaned web data, synthesized data, open-source resources, and in-house collected data. Its pre-training data covered a broad range of multimodal sources, including image captions, interleaved image--text data, OCR data, visual knowledge, multimodal academic questions, localization data, document parsing data, video descriptions, video localization, and agent interaction data. The model was therefore exposed to a much more diverse multimodal distribution than a single-image conversation dataset.

For this reason, calibrating Qwen2.5-VL with only the COCO-style conversational subset used in the LLaVA experiments would provide a calibration distribution that is too narrow relative to the capability scope of this backbone. Instead, we constructed a mixed-distribution calibration set that better reflects the broader multimodal coverage of Qwen2.5-VL. Specifically, the calibration set combined general image--text samples from LLaVA-Mix-665K~\cite{liu2024improved}, text-only samples from WikiText-2~\cite{merity2017pointer}, and multimodal samples from TextVQA~\cite{singh2019towards}, DocVQA~\cite{mathew2021docvqa}, and ChartQA~\cite{masry2022chartqa}. As a result, the final calibration set formed a broader mixed distribution spanning general multimodal conversation, pure text inputs, OCR-centric reasoning, document understanding, and chart understanding, providing a more suitable calibration distribution than a single-source setup.

\subsubsection{Candidate-layer selection.}
\begin{figure}[t]
  \centering
  \includegraphics[width=\linewidth]{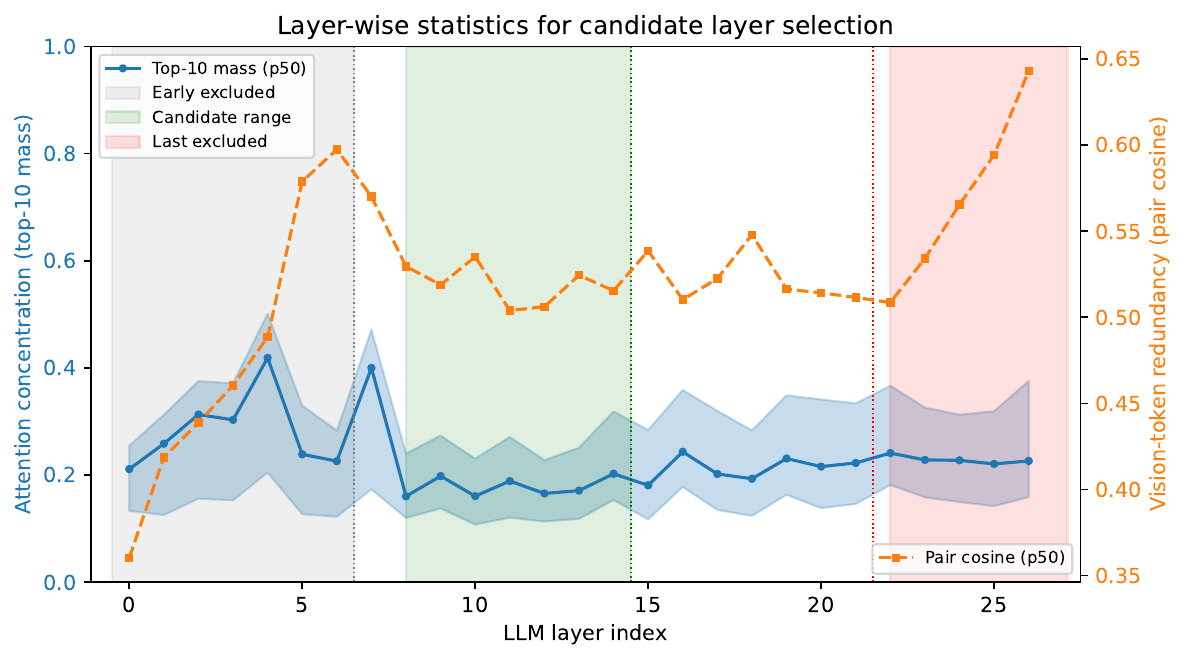}
  \caption{Candidate-layer selection on Qwen2.5-VL-7B. Layers 8--12 are used as the pruning candidate range.}
  \label{fig:qwen_layer_select}
\end{figure}
Following the same principle as in the main paper, we selected candidate pruning layers on Qwen2.5-VL-7B by profiling layer-wise attention concentration and vision-token redundancy on a calibration set. As shown in Fig.~\ref{fig:qwen_layer_select}, the earliest layers exhibited relatively weak and unstable attention concentration from text queries to visual tokens, making attention-based token ranking less reliable at shallow depth. In contrast, the later layers were closer to the final decision process and also showed increased vision-token redundancy, making pruning in these layers more sensitive. Based on these trends, we selected a contiguous middle range as the candidate set and used layers 8--12 in the Qwen2.5-VL-7B experiments. This choice followed the same design principle as in the main paper: avoid pruning too early, where cross-modal interaction is not yet stable, and avoid pruning too late, where token removal becomes more fragile.

\subsection{Main Results}

\begin{table}[!t]
\centering
\small
\setlength{\tabcolsep}{3.2pt}
\renewcommand{\arraystretch}{1.5}
\caption{Results on Qwen2.5-VL-7B. GFLOPs are reported using the same prefill-oriented estimation protocol as in the main experiments and are intended for approximate efficiency comparison.}
\resizebox{\linewidth}{!}{%
\begin{tabular}{l c c c c c c c c c c}
\toprule
\textbf{Method} & \textbf{Pre.} & \textbf{Keep Ratio} & \textbf{GFLOPs} & \textbf{MMB} & \textbf{MME} & \textbf{GQA} & \textbf{POPE} &  \textbf{SEED-I} & \textbf{SQA$^{I}$} & \textbf{Avg} \\
\midrule
\multirow{2}{*}{Original} & \multirow{2}{*}{FP16} & \multirow{2}{*}{100\%} & \multirow{2}{*}{100.00\%} & 83.68 & 2332.35 & 60.65 & 87.76 & 77.53 & 88.25 & \multirow{2}{*}{100.00\% } \\
  &  &  &  & 100.00\% & 100.00\% & 100.00\% & 100.00\% & 100.00\% & 100.00\% &  \\
\midrule
\multirow{2}{*}{\textbf{Pruning}} & \multirow{2}{*}{FP16} & \multirow{2}{*}{30\%} & \multirow{2}{*}{55.80\%} & 77.15 & 2278.27 & 49.30 & 87.90 & 74.76 & 84.13 & \multirow{2}{*}{93.85\%}  \\
  &  &  &  & 92.20\% & 97.68\% & 81.29\% & 100.16\% & 96.43\% & 95.33\% &  \\
VTW~\cite{lin2025boosting} & FP16 & 40\% & 43.43\% & 71.15\% & 66.82\% & 66.56\% & 70.37\% & -- & 90.87\% & 73.15\% \\
PDrop~\cite{xing2024pyramiddrop} & FP16 & 30\% & 33.03\% & 85.58\% & 86.50\% &82.62\% & 87.87\%  & -- & 94.65\% & 87.44\% \\
FastV~\cite{chen2024image}    & FP16 & 30\% & 32.66\% & 88.97\% & 94.36\% & 87.09\% & 96.00\% & -- & 96.48\% & 92.58\% \\
DivPrune~\cite{alvar2025divprune} & FP16 & 25\% & 24.45\% & 93.21\% & 96.95\% & 82.95\% & 97.71\% & -- & 95.18\% & 93.20\% \\
\midrule

\midrule
\multirow{2}{*}{\textbf{Quant.}} & \multirow{2}{*}{W4A4} & \multirow{2}{*}{100\%} & \multirow{2}{*}{100.00\%} & 79.81 & 2236.61 & 55.65 & 87.41 & 75.94 & 86.07 & \multirow{2}{*}{96.35\% } \\
  &  &  &  & 95.38\% & 95.90\% & 91.76\% &99.60\% & 97.95\% & 97.53\%  &  \\

GPTQ~\cite{frantar2023gptq}  & W4A16 & 100\% & 100.00\% & 97.09\% & 96.27\% & 99.34\% & 98.49\% & -- & -- & 97.80\% \\
AWQ~\cite{lin2024awq}   & W4A16 & 100\% & 100.00\% & 97.94\% & 97.00\% & 100.00\% & 99.42\% & -- & -- & 98.59\% \\
LUQ~\cite{bhatnagar2025luq}   & W2.75A16 & 100\% & 100.00\% & 77.12\% & 95.93\% & 87.44\% & 98.37\% & -- & -- & 89.72\% \\

\midrule

\midrule
\multirow{2}{*}{\textbf{Ours}} & \multirow{2}{*}{W4A4} & \multirow{2}{*}{30\%} & \multirow{2}{*}{55.80\%} & 78.01 & 2204.52 & 47.77 & 87.70 & 75.19 & 85.47 & \multirow{2}{*}{93.38\% } \\
  &  &  &  & 93.22\% & 94.52\% & 78.76\% & 99.93\% & 96.98\% & 96.85\% &  \\
\bottomrule
\end{tabular}
}
\label{tab:qwen_results}
\end{table}

Table~\ref{tab:qwen_results} reports the results on Qwen2.5-VL-7B under full precision, pruning-only, quantization-only, and the proposed collaborative setting. Overall, the collaborative pipeline remained effective on this backbone and preserved competitive performance under the W4A4 deployment setting, indicating that the main idea of calibration-aligned pruning is not limited to the LLaVA-style architecture.

Compared with the full-precision baseline, pruning-only with a 30\% keep ratio retained 93.85\% average performance, while quantization-only under W4A4 retained 96.35\%. The proposed collaborative setting reached 93.38\% average retention. Although the average result was slightly lower than the pruning-only and quantization-only reference settings, the collaborative model remained competitive and even surpassed the pruning-only baseline on several individual metrics, including MMB, SEED-I, and SQA$^{I}$. These results suggest that the calibration-derived pruning plan still transferred effectively to Qwen2.5-VL-7B and was able to preserve useful task-relevant tokens under low-bit execution.

The reported GFLOPs should also be interpreted with caution. Our GFLOPs followed the same prefill-oriented estimation protocol used in the main paper, but many prior methods did not disclose fully aligned FLOPs accounting details, and the final values can be noticeably affected by text token length. Therefore, the GFLOPs in Table~\ref{tab:qwen_results} are most useful for trend-level comparison under a consistent protocol, rather than as strict cross-method efficiency numbers. Under this interpretation, the collaborative setting achieved the same reported GFLOPs ratio as the pruning-only reference because both used the same final keep ratio and the same calibration-derived pruning plan, while the key difference lay in low-bit execution and score construction.

\subsubsection{Real Runtime Validation of the Quantized Path}
We further evaluate the selected quantized low-bit path under true INT4 execution to examine its practical deployment behavior. To this end, we instantiate a real quantized runtime with custom CUDA kernels and report per-task score together with deployment-oriented system metrics on both LLaVA-1.5-7B and Qwen2.5-VL-7B.

\begin{table*}[t]
\centering
\small
\caption{Per-task real-runtime measurements under true INT4 execution on LLaVA-1.5-7B and Qwen2.5-VL-7B.}
\setlength{\tabcolsep}{3.5pt}
\renewcommand{\arraystretch}{1.15}
\resizebox{\textwidth}{!}{%
\begin{tabular}{c|cccccc|cccccc}
\toprule
\multirow{2}{*}{\textbf{Task}} &
\multicolumn{6}{c|}{\textbf{LLaVA-1.5-7B} (\textbf{14.13\,GB FP16 / 3.97\,GB INT4})} &
\multicolumn{6}{c}{\textbf{Qwen2.5-VL-7B} (\textbf{16.6\,GB FP16 / 5.86\,GB INT4})} \\
\cmidrule(lr){2-7} \cmidrule(l){8-13}
& \multicolumn{2}{c}{\textbf{ACC.}}
& \multicolumn{2}{c}{\textbf{Peak GPU Mem}}
& \multicolumn{2}{c|}{\textbf{Avg GPU Power}}
& \multicolumn{2}{c}{\textbf{ACC.}}
& \multicolumn{2}{c}{\textbf{Peak GPU Mem}}
& \multicolumn{2}{c|}{\textbf{Avg GPU Power}} \\
\cmidrule(lr){2-3} \cmidrule(lr){4-5} \cmidrule(lr){6-7}
\cmidrule(lr){8-9} \cmidrule(lr){10-11} \cmidrule(l){12-13}
& \textbf{FP16} & \textbf{INT4} & \textbf{FP16} & \textbf{INT4} & \textbf{FP16} & \textbf{INT4}
& \textbf{FP16} & \textbf{INT4} & \textbf{FP16} & \textbf{INT4} & \textbf{FP16} & \textbf{INT4} \\
\midrule
MMBench & 62.37 & 57.90 & 15.59 & 6.12 & 384.77 & 195.39 & 83.68 & 81.87 & 17.36 & 7.06 & 329.46 & 209.60 \\
SEED-I  & 63.94 & 62.21 & 15.48 & 6.17 & 314.77 & 95.76  & 77.53 & 77.46 & 17.78 & 7.39 & 347.64 & 293.21 \\
SQA-I   & 66.24 & 63.16 & 15.74 & 6.27 & 315.45 & 221.89 & 88.25 & 86.56 & 17.48 & 7.43 & 322.56 & 180.72 \\
MME     & 1767.46 & 1714.72 & 14.75 & 5.45 & 324.37 & 196.80 & 2332.35 & 2265.62 & 17.73 & 7.33 & 316.69 & 251.21 \\
POPE    & 85.59 & 85.42 & 14.75 & 5.45 & 385.00 & 221.08 & 87.76 & 86.84 & 17.47 & 7.25 & 331.91 & 218.30 \\
GQA     & 60.65 & 59.81 & 15.52 & 6.22 & 345.10 & 194.31 & 60.65 & 59.41 & 17.44 & 7.25 & 325.62 & 213.43 \\
\bottomrule
\end{tabular}
}
\label{tab:real_runtime_combined}
\end{table*}

Table~\ref{tab:real_runtime_combined} shows that real INT4 quantization materially compresses both backbones, reducing the on-disk model size from 14.13\,GB to 3.97\,GB on LLaVA-1.5-7B and from 16.6\,GB to 5.86\,GB on Qwen2.5-VL-7B. Meanwhile, peak GPU memory and average GPU power are reduced consistently across all evaluated tasks. The real-quantized results also remain broadly consistent with the fake-quant evaluation, indicating that the selected low-bit regime transfers reliably from calibration-time analysis to actual deployment. Overall, these measurements support the practical deployability of the selected quantized path and reinforce the deployment relevance of the low-bit regime adopted in this work.